\documentclass[preprint]{article}

\usepackage{neurips_2024}

\usepackage[utf8]{inputenc} %
\usepackage[T1]{fontenc}    %
\usepackage{hyperref}       %
\usepackage{url}            %
\usepackage{booktabs}       %
\usepackage{amsfonts}       %
\usepackage{nicefrac}       %
\usepackage{microtype}      %
\usepackage{xcolor}         %
\usepackage{graphicx}
\usepackage{amsmath} %
\usepackage{amssymb}  %
\usepackage{amsfonts}
\usepackage{mathtools}
\usepackage{commath}
\usepackage{color}
\usepackage{multirow}
\usepackage{makecell} %

\usepackage{caption}

\usepackage{pifont}   %
\newcommand{\cmark}{\ding{51}}  %
\newcommand{\xmark}{\ding{55}}  %

\title{OccSora: 4D Occupancy Generation Models as \\ World Simulators for Autonomous Driving}

\begin{document}

\newcommand*\samethanks[1][\value{footnote}]{\footnotemark[#1]}
\DeclareUrlCommand\url{\color{magenta}}

\author{Lening Wang$^{1,2,}$\footnotemark[1] $^,$\footnotemark[2] \quad 
Wenzhao Zheng$^{2,3,}$\footnotemark[1] $^,$\footnotemark[3]  \quad 
Yilong Ren$^1$\quad
Han Jiang$^{1}$ \\ \quad
\textbf{Zhiyong Cui}$^{1}$\quad 
\textbf{Haiyang Yu}$^{1}$\quad 
\textbf{Jiwen Lu}$^{3}$ \\
\url{https://wzzheng.net/OccSora}\\
$^1$State Key Lab of Intelligent Transportation System, Beihang University, China \\
$^2$EECS, UC Berkeley, United States \quad\quad $^3$Department of Automation, Tsinghua University, China \\
\texttt{leningwang@buaa.edu.cn; wenzhao.zheng@outlook.com}
}
\maketitle

\renewcommand{\thefootnote}{\fnsymbol{footnote}}
\footnotetext[1]{Equal contribution. $\dagger$Work done during an internship at UC Berkeley. $\ddagger$Project leader. 
}
\renewcommand{\thefootnote}{\arabic{footnote}}

\begin{abstract}

Understanding the evolution of 3D scenes is important for effective autonomous driving.
While conventional methods model scene development with the motion of individual instances, world models emerge as a generative framework to describe the general scene dynamics.
However, most existing methods adopt an autoregressive framework to perform next-token prediction, which suffer from inefficiency in modeling long-term temporal evolutions.
To address this, we propose a diffusion-based 4D occupancy generation model, OccSora, to simulate the development of the 3D world for autonomous driving.
We employ a 4D scene tokenizer to obtain compact discrete spatial-temporal representations for 4D occupancy input and achieve high-quality reconstruction for long-sequence occupancy videos.
We then learn a diffusion transformer on the spatial-temporal representations and generate 4D occupancy conditioned on a trajectory prompt. 
We conduct extensive experiments on the widely used nuScenes dataset with Occ3D occupancy annotations.
OccSora can generate 16s-videos with authentic 3D layout and temporal consistency, demonstrating its ability to understand the spatial and temporal distributions of driving scenes.
With trajectory-aware 4D generation, OccSora has the potential to serve as a world simulator for the decision-making of autonomous driving.
Code is available at: \url{https://github.com/wzzheng/OccSora.}
\end{abstract}

\section{Introduction}

As a promising application of artificial intelligence technology, autonomous driving has garnered widespread attention and research in recent years \cite{hu2023planning, fu2024drive,yang2023bevformer}. 
Establishing the relationship between perception \cite{liu2023bevfusion, chang2023bev,chen2023futr3d,mao20233d}, prediction~\cite{hu2021fiery,gu2022vip3d,liang2020pnpnet}, and planning \cite{mozaffari2020deep, huang2023differentiable,jia2023think,wang2024deepaccident} in autonomous driving is crucial for a comprehensive understanding of the field. 

Conventional autonomous driving models \cite{hu2023planning} rely on the motion of the ego vehicle instances to model the development of scenes, unable to develop a profound understanding of scene perception and vehicle motion control comparable to human understanding. 
The emergence of world models \cite{World2018} offers new possibilities for a deeper understanding of the comprehensive relationship between autonomous driving scenes and vehicle motion.
Based on strong image pretrained models, image-based world models~\cite{hu2023gaia, wang2023drivedreamer} can generate high-quality driving-scene images with conditions of 3D bounding boxes.
OccWorld~\cite{zheng2023occworld} further learns a world model in the 3D occupancy space, which can be better leveraged for 3D reasoning for autonomous driving. However, most existing methods adopt an autoregressive framework to model the dynamics (e.g., image tokens, bounding boxes, occupancy) of a 3D scene, which hinders their ability to efficiently produce long-term video sequences.

\begin{figure}[t]
  \centering
  \includegraphics[width=1\linewidth]{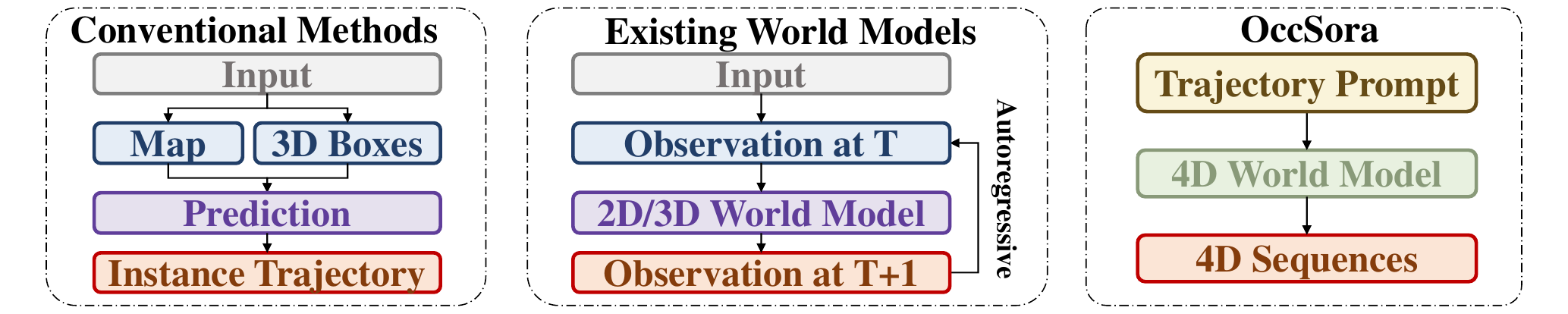}
\vspace{-6mm} 
  \caption{\textbf{Comparisons with existing methods.} It can comprehend the intricate relationship between scenes and trajectories and generate long-term, physically consistent 4D occupancy.}
  \label{fig0}
  \vspace{-7mm} 
\end{figure}

To address this, we propose a 4D world model OccSora to directly generate spatial-temporal representations with diffusion models as shown in Figure~\ref{fig0}, motivated by OpenAI's 2D video generation model Sora~\cite{videoworldsimulators2024}.
To accurately understand and represent 4D scenes, we design 4D scene discretization to capture the dynamic characteristics of scenes and propose a diffusion-based world model to achieve controllable scene generation following physical laws. Specifically, in the 4D occupancy scene tokenizer, we focus on extracting and compressing real 4D scenes to establish an understanding of the world model environment. In the diffusion-based world model, we employ multidimensional diffusion techniques to propagate accurate spatiotemporal 4D information and realize trajectory-controllable scene generation by incorporating real ego car trajectories as condition, thereby achieving a deeper understanding between autonomous driving scenes and vehicle motion control. Through training and testing, OccSora can generate autonomous driving 4D occupancy scenes that adhere to physical logic and achieve controllable scene generation based on different trajectories. The proposed autonomous driving 4D world model opens up new possibilities for understanding dynamic scene changes in autonomous driving and the physical world.

\section{Related Work}

\textbf{3D Occupancy Prediction.} 3D occupancy focuses on partitioning space into voxels and assigning specific semantic types to each voxel. It is considered a crucial means of representing real-world scenes, following 3D object detection \cite{mao20233d,ma20233d,yu2024flow} and Bird's Eye View (BEV) perception \cite{yang2023bevheight,zhao2024bev,wang2023bevgpt,zhang2022beverse}, for autonomous driving perception tasks. Early research on this task primarily focused on semantically classifying discrete points from LiDAR \cite{zhou2021panoptic,singh2020range,liu2023lidar,zuo2023pointocc}. In fact, due to the camera containing semantic information far exceeding that of LiDAR and their low cost. Thus, utilizing images for depth estimation or employing end-to-end methods for 3D scene perception research is currently the mainstream approach \cite{huang2023tri, li2023voxformer, wei2023surroundocc, huang2023selfocc}. Considering the advantages of multi-sensor systems, some studies research multi-modal fusion for 3D occupancy prediction \cite{wang2023openoccupancy,zhang2024occfusion}.

In addition to utilizing typical sensor devices for 3D occupancy prediction, some studies focus on other tasks involving occupancy. For instance, OccWorld \cite{zheng2023occworld} proposes a spatiotemporal generative transformer to predict subsequent scene tokens and the vehicle token, thereby predicting future occupancy and vehicle trajectory. On the other hand, GenOcc \cite{wang2024occgen} utilizes generative models to accomplish occupancy prediction. DriveWorld \cite{min2024driveworld} introduces a world-model-based framework for learning in autonomous driving from 2D images and videos, addressing tasks such as 3D object detection, online map creation, and occupancy prediction. 
Although progress has been made in 3D occupancy prediction and continuous 4D prediction, the scope of these studies remains limited. 
They usually use autoregressive models in conjunction with scene information from preceding frames to carry out subsequent occupancy tasks, thereby necessitating prior scene or 3D bounding box inputs. Consequently, they lack a genuine understanding of the fundamental relationships between scene and motion, and therefore do not constitute world models conditioned on actions.

\begin{figure}[t]

  \centering
  \includegraphics[width=1\linewidth]{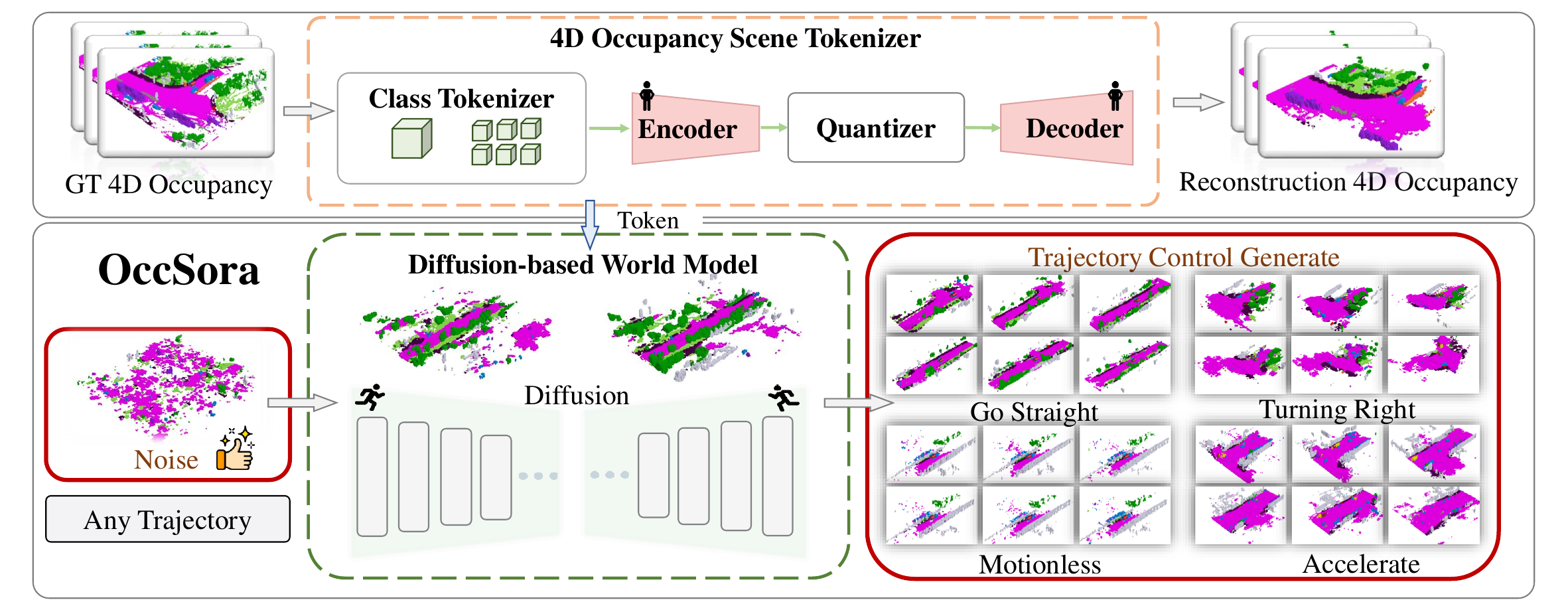}
\vspace{-5mm} 
  \caption{\textbf{The pipeline of OccSora.} The 4D occupancy scene tokenizer achieves compression and restoration of real information. The compressed information and vehicle trajectories are simultaneously used as inputs for the diffusion-based world model. After training, the diffusion-based world model utilizes random noise and arbitrary trajectories to generate controllable tokens, which are then decoded into 4D occupancy maps in the 4D occupancy scene tokenizer stage. }
  \label{fig1}
 \vspace{-7mm}
\end{figure}

\textbf{Generative Model.} 
Generative models have garnered widespread attention recently due to their powerful capabilities. 
By learning the probability distribution of data, generative models can train models capable of generating new samples. From the emergence of Generative Adversarial Networks (GAN) \cite{goodfellow2020generative} to the recent advent of diffusion models like Variational Autoencoders (VAE) \cite{van2017neural}, the tasks of generative models have gradually expanded from initial image generation tasks to in-depth studies on videos \cite{yan2021videogpt}. Tasks such as image generation based on the DIT model \cite{peebles2023scalable} delve into and utilize their generative capabilities. The Sora video generation model \cite{videoworldsimulators2024} further demonstrates the ability to produce high-quality videos with realistic transitions between frames in continuous scenes. 

Similarly, in the field of autonomous driving, controllable image generation can provide various driving scenarios to serve perception, planning, control, and decision-making tasks. For instance, MagicDriver \cite{gao2023magicdrive} generates videos depicting various weather scenarios by learning from videos of autonomous driving vehicles and incorporating labels such as object detection boxes and maps. DriveDreamer \cite{wang2023drivedreamer} proposes a world model that is entirely derived from real-world driving scenes, enabling a deep understanding of structured traffic constraints and thereby achieving precise and controllable video generation. However, for autonomous driving scenarios, obtaining the 3D occupancy of scenes is more important compared to 2D information \cite{zhang2023occformer, mescheder2019occupancy, sima2023_occnet}. Some studies \cite{lee2024semcity, liu2023pyramid} propose a three-dimensional diffusion model suitable for generating outdoor real scenes, which, by utilizing diffusion methods, accomplishes scalable seamless scene generation tasks. While some previous studies have generated 2D static images and extended them to the temporal dimension through autoregression, and others have achieved static generation of 3D occupancy scenes, both the 2D images generated based on 3D object bounding boxes and the static large-scale scenes are difficult to directly apply to autonomous driving tasks \cite{wang2024deepaccident, zheng2024genad}. In contrast, our proposed OccSora establishes a dynamic 4D occupancy world model that adapts to scene changes with vehicle trajectories, without the need for any prior object detection boxes or scene information, representing the first generative 4D occupancy world model for autonomous driving.

\section{Proposed Approach}

\subsection{World Model for Autonomous Driving} 
4D occupancy can comprehensively capture the structural, semantic, and temporal information of a 3D scene and effectively facilitate weak supervision or self-supervised learning, which can be applied to visual, LiDAR, or multimodal tasks. 
Based on these principles, we represent the world model $\chi$ as 4D occupancy $R$. Figure \ref{fig1} illustrates the overall framework of OccSora. We constructed a 4D occupancy scene tokenizer to compress real 4D occupancy $R_{in} \in \mathbb{R}^{B\times D\times H\times W\times T}$ in both the temporal $T$ and spatial $D\times H\times W$ dimensions, capturing the relationships and evolution patterns in 4D autonomous driving scenes. This results in compressed high-level tokens $R_{mi} \in \mathbb{R}^{B\times c\times h\times w\times t}$ and reconstructed 4D occupancy data $R_o \in \mathbb{R}^{B\times D\times H\times W\times T}$. We designed a diffusion-based world model that uses trajectory information $R_{tr} \in \mathbb{R}^{B\times T\times 2}$ as control units, training them supervised by the compressed tokens $R_{mi}$ to generate high-dimensional scene representation tokens $T_o \in \mathbb{R}^{B\times c\times h\times w\times t}$. They are then decoded by the 4D occupancy scene tokenizer to consistent and dynamically controllable $R_o$.

\subsection{4D Occupancy Scene Tokenizer}

\begin{figure}[t]
  \centering
  \includegraphics[width=1\linewidth]{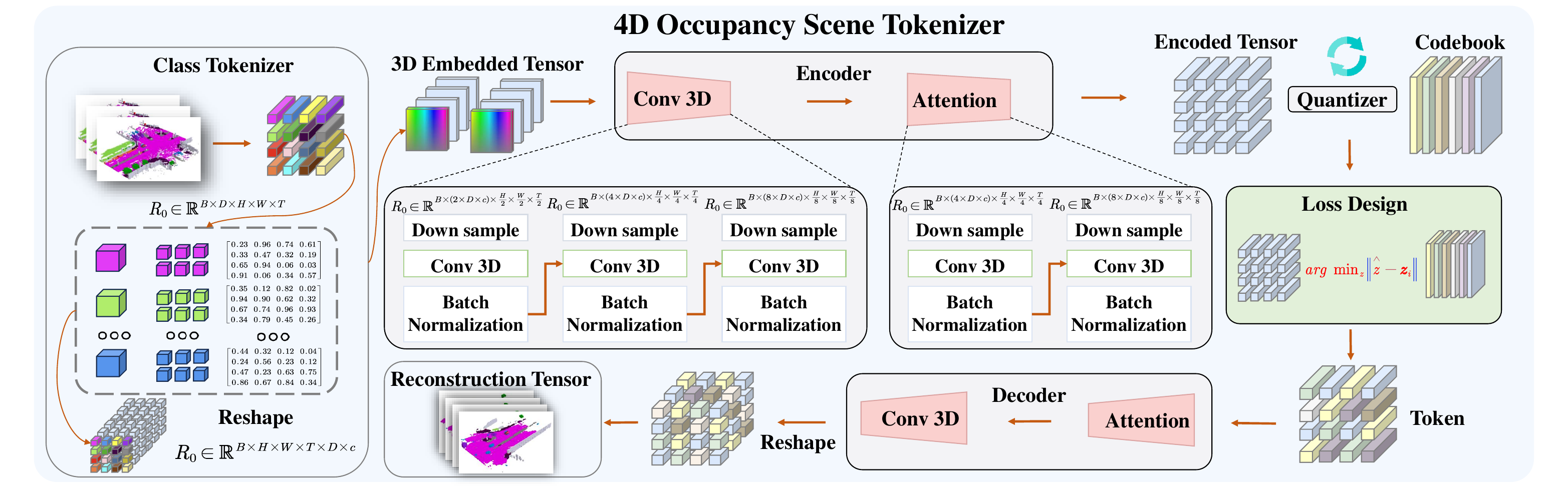}
\vspace{-5mm} 
  \caption{\textbf{The structure of the 4D occupancy scene tokenizer.} The proposed method encodes and compresses 4D scenes to extract high-dimensional features, which are then decoded to retrieve the spatiotemporal physical characteristics of the scenes.}
  \label{fig2}
\vspace{-5mm} 
\end{figure}

The goal of 4D occupancy prediction is to determine the semantic type at specific locations over time. We discretize and encode the real 4D occupancy scene $R_{in} $ into an intermediate latent space $R_{mi}$ to obtain a true representation of the 4D occupancy scene, as shown in Figure \ref{fig2}. The formula is as follows: $
    R_{mi}=\zeta _{token}\left\{ \tau _{en}\left( R_{in} \right) \right\}.$ Here, $\zeta_{token}$ represents the encoded codebook, and $\tau_{en}$ denotes the designed 3D encoder network and category embedding. This 3D occupancy representation divides the 3D space around the vehicle into voxels $r^T=N \in \mathbb{R}^{H \times W \times D}$, where each voxel position is assigned a type label $N$, indicating whether it is occupied and the semantics of the object occupying it. Unlike traditional methods, we incorporate and compress temporal information within the same scene, reshaping the tensor to $R_{in}$. This approach allows for unified learning of both spatial and temporal evolution patterns and the physical relationships of real scenes, compared to previous autoregressive methods. After passing through the $\tau_{en}$ 3D encoder network with category embedding and the $\zeta_{token}$ encoded codebook, the tensor is transformed into $R_{mi}$ represents the potential spaces. This reshaping ensures a comprehensive representation of the temporal dynamics of 4D occupancy.

\textbf{Category Embedding and Tokenizer.} To accurately capture the spatial information of the original parameters, we first perform an embedding operation on the input $R_{in}$. We assign a learnable category embedding \( b \in \mathbb{R}^{c'} \) for each category in \( R_{in} \) to label the categories of continuous 3D occupancy scenes. The position information is embedded as tokens that represent the categories. Then, these embeddings are concatenated along the feature dimension. To facilitate subsequent 3D encoding with compression in specific dimensions, we further reshape \( R_{in} \) into \( R_{in}' \in \mathbb{R}^{B \times (Dc') \times T \times H \times W} \).

\textbf{3D Video Encoder.} To effectively learn discrete latent tokens, we further performed downsampling on the embedded positional information of the 4D occupancy $R_{in}'$ to extract high-dimensional features. The designed encoder architecture comprises a series of 3D downsampling convolutional layers, which perform 3D downsampling in both the time dimension (T) and spatial dimensions (H × W), increasing the fusion dimension to $D \times c'$. We initially downscaled the input $R_{in}'$ three times to obtain $R_{in}''\in \mathbb{R}^{B \times (8 \times Dc') \times \frac{T}{8} \times \frac{H}{8} \times \frac{W}{8}}$, and introduced dropout layers after the feedforward and attention block layers for regularization. Considering the relationships between consecutive frames, we introduced cross-channel attention after downsampling, segmenting $R_{in}''$ along the $8 \times Dc'$ dimension and then performing cross-channel attention between the segmented parts. This operation enhanced the model ability to capture relationships between features along different axes, and subsequently reshaped them back to the original shape to obtain the output tensor \( R_{mi} \).

\textbf{Coodbook and Training Objective.} To achieve a more condensed representation, we simultaneously learn a codebook $\zeta _{token} \in \mathbb{R}^{N \times D}$ containing N codes. Each code $b \in \mathbb{R}^{c'}$ in the codebook encodes a high-level concept of the scene, such as whether the corresponding position is occupied by a car. $\zeta _{token}$ represents the encoded codebook. We quantize each spatial feature $\widehat{R_{mi}^{\left( ij \right)}}$ in $\widehat{R_{mi}^{}}$ by mapping it to the nearest code $N(\widehat{R_{mi}^{\left( ij \right)}}, B)$:\begin{equation}
R_{mi}^{\left( ij \right)} = N(\widehat{R_{mi}^{\left( ij \right)}}, \zeta _{token} ) = \min_{b \in \zeta _{token}} ||\widehat{R_{mi}^{\left( ij \right)}} - b||_2, 
\end{equation}where $|| \cdot ||_2$ represents the L2 norm. Subsequently, we integrate the quantized features $\widehat{R_{mi}^{\left( ij \right)}}$ to obtain the final scene representation $R_{mi}$.

\textbf{3D Video Decoder.} To reconstruct $R_o$ from the learned scene representation $R_{mi}$, we design a decoder consisting of 3D deconvolution layers. In contrast to the encoder, the decoder architecture includes cross-channel attention, residual blocks, and a series of 3D convolutions, enabling upsampling in both temporal and spatial dimensions. This gradual upsampling process transforms $
R_{mi} $ to its original occupancy resolution $R_o $. The decoder then splits the result along the channel dimension to reconstruct the temporal dimension, yielding occupancy values for each voxel. 
During training, we accomplished the training of the encoder, decoder parameters, and the encoding codebook. The designed network enables us to simultaneously encode the input 4D occupancy information and compress it into multiple tokens, thereby learning the physical correlations of world models under spatiotemporal fusion. 
Additionally, we restore the information during the decoding process.

\subsection{Diffusion-based World Model}

\begin{figure}[t]
  \centering
  \includegraphics[width=1\linewidth]{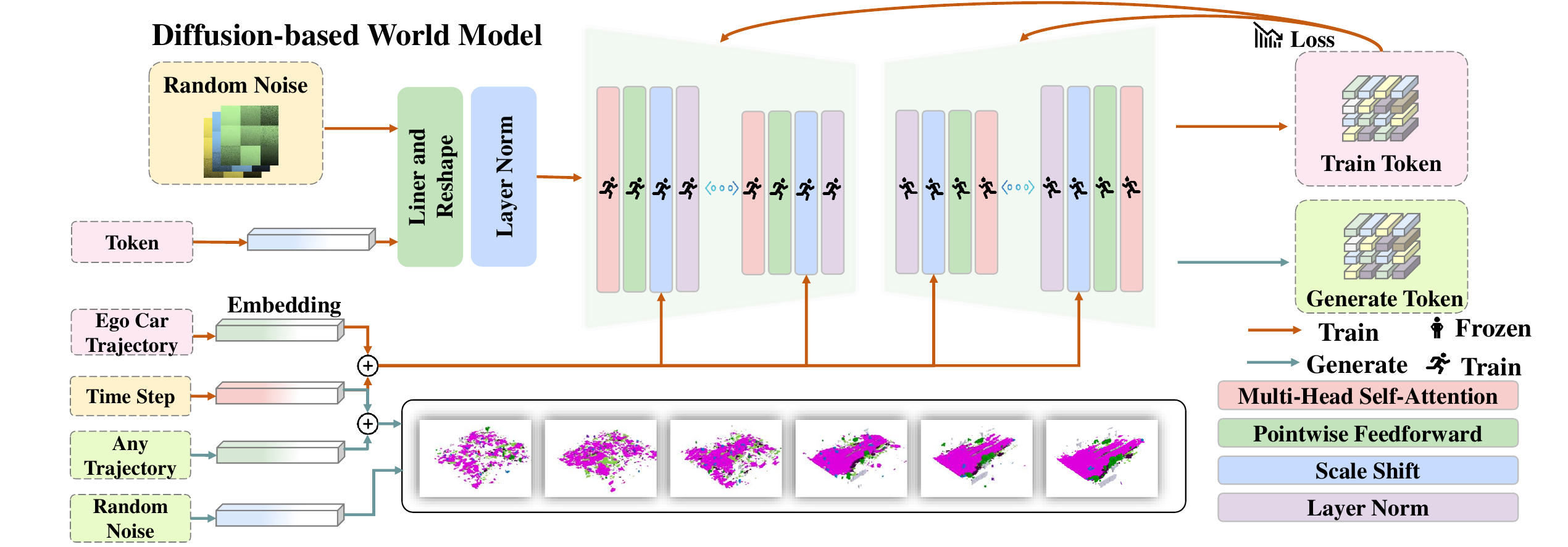}
\vspace{-5mm} 
  \caption{\textbf{The structure of the diffusion-based world model.} The model involves utilizing the optimal codebook obtained from training the 4D occupancy scene tokenizer to convert 4D occupancy into a sequence of tokens. These tokens, along with the ego vehicle trajectory and random noise, are then combined as input for denoising training to acquire the generated token.}
  \label{fig4}
\vspace{-5mm} 
\end{figure}

Inspired by the diffusion method \cite{Peebles2022DiT}, we use scene tokens $R_{mi}$ containing spatiotemporal information features as inputs for the generative model. Additionally, we conduct denoising training and trajectory-controllable generation tasks under the control of vehicle trajectories $R_{tr}$, as shown in Figure \ref{fig4}.

\textbf{Token Embedding.} To efficiently and accurately utilize the transformer \cite{vaswani2017attention}, we flatten the input data tokens $R_{mi}$ into $R_{\text{re}} $. Simultaneously, considering the significance of positional information for spatiotemporal compression, we perform positional embedding on the input. We design the following function, which utilizes sin and cos functions to encode positional indices: \begin{equation}
R_{re}^{\left( \mathrm{emb} \right)}=\mathrm{emb}_{\mathrm{i}}^{\mathrm{d}}+R_{re},R_{re} \in \mathbb{R} ^{\mathrm{B}\times \mathrm{c}\times \left( \mathrm{hwt} \right)}.
\end{equation} It operates on two main parameters: \( C \), representing the embedding output dimensionality of each position, and \( i = hwt \), representing the number of tokens enumerating the positions to be encoded. The resulting output follows a matrix structure of dimensions \( C \times i \), and \( emb \) constructs the positional embedding representation using sin and cos functions. These embeddings encapsulate the positional attributes of the tokens, enhancing the model understanding of positions within the input. We add the positional encoding \(E_{mb}\) to the input \(R_{re}\), yielding \(R_{re}^{\left( \mathrm{emb} \right)}\), which represents the tokens after positional encoding.

\textbf{Trajectory Conditioning Embedding.} The transformation relationship between scenes and trajectories is a crucial aspect of autonomous driving. Generating diverse 4D occupancy scenes that align with control trajectories is essential. Therefore, we use the ego vehicle trajectory \(T_r\) as input to generate controllable 4D occupancy. Firstly, the ego vehicle trajectory \(T_r \in \mathbb{R}^{B\times t\times 2}\) is used as one of the control inputs, where \(t\) denotes the continuous time dimension, and the third dimension represents the vehicle positions along the $x$ and $y$ axes of the absolute coordinate system. To achieve trajectory embedding and encoding, we reshape the vehicle trajectory to \(T_r \in \mathbb{R}^{B\times \left( t\times 2 \right)}\) and learn and encode it as follows: \begin{equation}
g=\nu\left( t \right) +\delta \left( T_r \right) ,\delta \left( T_r \right) \in \mathbb{R} ^{\mathrm{B}\times \mathrm{c}\times \left( \mathrm{hwt} \right)},\nu \in \mathbb{R} ^{\mathrm{B}\times \mathrm{c}\times \left( \mathrm{hwt} \right)},
\end{equation} where $\nu \in \mathbb{R} ^{\mathrm{B}\times \mathrm{c}\times \left( \mathrm{hwt} \right)}
$ represents the time step embedding, and \(\delta\) denotes the Multilayer Perceptron (MLP) network that extracts trajectory information. It is then embedded $g$ into the input sequence of the diffusion transformer and processed together with the token information $R_{re}^{\left( \mathrm{emb} \right)}$.

\textbf{Diffusion Transformer.} We developed a diffusion-based world model to learn from and generate within the latent space \(R_{mi}\), while integrating trajectory labels \(T_r\) and denoising time steps \(\nu\) as control conditions. In the model diffusion learning process, we constructed a forward noise process that gradually introduces noise to the latent space \(R_{mi}\): 
$
    q\left( R_{\mathrm{re}}^{g}|R_{\mathrm{re}} \right) ={N} \left( R_{\mathrm{re}}^{g};\sqrt{\overline{\sigma ^g}}R_{\mathrm{re}},\left( 1-\overline{\sigma ^g} \right) I \right) ,
$  where the constant \(g\) represents the embedding of trajectories and time steps. Utilizing the reparameterization trick, we can sample: $
R_{\mathrm{re}}^{g}=\sqrt{\overline{\sigma ^g}}R_{\mathrm{re}}+\sqrt{1-\overline{\sigma ^g}}\epsilon ^g$, where \(\epsilon^g \sim {N} \left( 0, I \right)\). The 4D occupancy diffusion model is trained to learn the reverse propagation process. To invert the forward process corruption: 
\begin{equation}
p_{\theta}\left( R_{\mathrm{re}}^{g-1}|R_{\mathrm{re}}^{g} \right) ={N} \left( \mu _{\theta}\left( R_{\mathrm{re}}^{g} \right) ,\Sigma_{\theta}\left( R_{\mathrm{re}}^{g} \right) \right) ,
\end{equation}
 where neural networks predict the statistical properties of \(p\). The reverse process model is trained with the variational lower bound of \(x_0\), which simplifies to: $
{L} \left( \theta \right) =-p\left( R_{\mathrm{re}}^{0}|R_{\mathrm{re}}^{1} \right) +\sum{_g{D} _{KL}\left( q^*\left( R_{\mathrm{re}}^{g-1}|R_{\mathrm{re}}^{g},R_{\mathrm{re}}^{0} \right) \right) ||p_{\theta}\left( R_{\mathrm{re}}^{g-1}|R_{\mathrm{re}}^{g} \right)},
$ which excluding irrelevant additional terms during training. As both \(q\) and \(p\) are gaussian distributions, the Kullback-Leibler (KL) divergence can be evaluated using the means and covariances of the two distributions. By reparameterizing as a noise prediction network, the model can be trained using the simple mean squared error between the predicted noise \(\hat{R}_{re}^{g}\) and the sampled gaussian noise \(R_{re}^{g}\): $
{L} _{simple}(\theta )=\frac{1}{2}(\hat{R}_{re}^{g}-R_{re}^{g})^2 .
$ However, to train the diffusion model with learned reverse process covariance, the full KL divergence term needs to be optimized. We follow diffusion models approach \cite{dhariwal2021diffusion}: train first with \({L}_{simple}(\theta )\), then with the full \({L}\). Once \(p\) is trained, new token can be sampled by initializing \(R_{re}^g \sim N(0,I)\) and sampling \(R_{re}^{g-1} \sim p(R_{re}^{g-1}|R_{re}^g)\) using the reparameterization trick.

Overall, tokens \(R_{mi}\) processed in the initial stage as \(R_{re}\) are passed to a series of transformer blocks for further refinement. These blocks effectively capture the relationships between trajectory information and tokens. Regarding noisy image input processing, the diffusion transformer employs specific attention mechanisms and loss functions to minimize the impact of noise on model performance, ensuring robust operation in noisy environments. To incorporate trajectory labels \(T_r\) and denoising time steps \(\nu\) as additional control conditions, we feed them as supplementary inputs alongside token embeddings into the transformer blocks. This enables the model to dynamically adjust its processing based on these conditions, thereby better adapting to various trajectory control requirements. In the end, the trained diffusion-based world model successfully transforms pure noise and trajectory labels \(T_r\) into \(T_o \in \mathbb{R}^{B\times c\times h\times w\times t}\), which are eventually decoded into \(R_{o}\) through the 3D decoder.

\section{Experiments}

As a 4D occupancy world model in the field of autonomous driving, OccSora offers a deeper understanding of the relationship between autonomous driving scenes and vehicle trajectories without requiring any input of 3D bounding boxes, maps, or historical information. It can construct a long-time sequence world model that adheres to physical laws. We have conducted a series of quantitative experiments and visualizations to illustrate this.

\subsection{Implementation Details} 

We conducted experiments on the widely used nuScenes-Occupancy dataset \cite{caesar2020nuscenes}, which is currently one of the most mainstream and standard datasets, supporting many well-known research studies \cite{hu2023planning,wang2023openoccupancy}. For the OccSora model, we applied three rounds of compression to 32 consecutive frames and increased its channel dimension to 128. Subsequently, we conducted further comparative and ablation experiments under different components and trajectory scenarios. We trained using the AdamW optimizer with an initial learning rate set to \(1 \times 10^{-5}\) and a weight decay of 0.01. Using 8 NVIDIA GeForce A100 GPUs, we set a batch size of 2 per GPU. For the training of the 4D occupancy scene tokenizer, we needed about 42GB of memory per GPU to train for 150 epochs, which took 50.6 hours. For the diffusion-based world model, we needed about 47GB of memory per GPU to train for 1,200,000 steps, which took 108 hours.

\begin{table*}[t!]
\small 
\caption{\textbf{The quantitative analysis of 4D occupancy reconstruction.} Despite a compression rate 32 times greater than OccWorld, OccSora maintains over half of its reconstruction accuracy compared to OccWorld.}
\centering
\setlength{\tabcolsep}{0.5pt}
\vspace{-2mm}
\begin{tabular}{@{}l|c|cc|ccccccccccccccccc@{}}
\Xhline{1pt} 
Method &Ratio&IoU&mIoU& {\rotatebox[origin=c]{90}{  Others  }} & {\rotatebox[origin=c]{90}{  barrier  }}& {\rotatebox[origin=c]{90}{  bicycle  }}& {\rotatebox[origin=c]{90}{  bus  }}& {\rotatebox[origin=c]{90}{  car  }}& {\rotatebox[origin=c]{90}{  const. veh.  }}& {\rotatebox[origin=c]{90}{  motorcycle  }}& {\rotatebox[origin=c]{90}{  pedestrian  }}& {\rotatebox[origin=c]{90}{  traffic cone  }}& {\rotatebox[origin=c]{90}{  trailer  }}& {\rotatebox[origin=c]{90}{  truck  }}& {\rotatebox[origin=c]{90}{  drive. suf.  }}& {\rotatebox[origin=c]{90}{  other flat  }} & {\rotatebox[origin=c]{90}{  sidewalk  }}& {\rotatebox[origin=c]{90}{  terrain  }}& {\rotatebox[origin=c]{90}{  man made  }} & {\rotatebox[origin=c]{90}{  vegetation  }} \\
& &&& \rotatebox[origin=c]{90}{\colorbox{black}{}} & \rotatebox[origin=c]{90}{\colorbox[RGB]{255,120,50}{}} & \rotatebox[origin=c]{90}{\colorbox[RGB]{255,192,203}{}} & \rotatebox[origin=c]{90}{\colorbox[RGB]{252,254,88}{}} & \rotatebox[origin=c]{90}{\colorbox[RGB]{87,149,237}{}} & \rotatebox[origin=c]{90}{\colorbox[RGB]{140,251,253}{}} & \rotatebox[origin=c]{90}{\colorbox[RGB]{193,180,61}{}} & \rotatebox[origin=c]{90}{\colorbox[RGB]{222,51,35}{}} & \rotatebox[origin=c]{90}{\colorbox[RGB]{249,240,162}{}} & \rotatebox[origin=c]{90}{\colorbox[RGB]{120,64,21}{}} & \rotatebox[origin=c]{90}{\colorbox[RGB]{145,52,231}{}} & \rotatebox[origin=c]{90}{\colorbox[RGB]{226,59,246}{}} & \rotatebox[origin=c]{90}{\colorbox[RGB]{138,137,137}{}} & \rotatebox[origin=c]{90}{\colorbox[RGB]{65,10,72}{}} & \rotatebox[origin=c]{90}{\colorbox[RGB]{176,237,107}{}} & \rotatebox[origin=c]{90}{\colorbox[RGB]{83,112,152}{}} & \rotatebox[origin=c]{90}{\colorbox[RGB]{90,172,52}{}}\\
\midrule
        OccWorld \cite{zheng2023occworld} & 16& 62.2& 65.7 & 45.0 & 72.2 & 69.6 & 68.2 & 69.4 & 44.4 & 70.7 & 74.8 & 67.6 & 54.1 & 65.4 & 82.7 & 78.4 & 69.7 & 66.4 & 52.8 & 43.7\\
\hline
        OccSora &512&37.0&27.4&11.7&22.6 & 0.0 & 34.6 & 29.0 & 16.6 & 8.7 & 11.5 & 3.5 & 20.1 & 29.0 & 61.3 & 38.7 & 36.5 & 31.1 & 12.0 & 18.4\\

\Xhline{1pt} 
\end{tabular}
\vspace{-3mm}
\label{tab1}
\end{table*}

\begin{figure}[t!]
  \centering
  \includegraphics[width=1\linewidth]{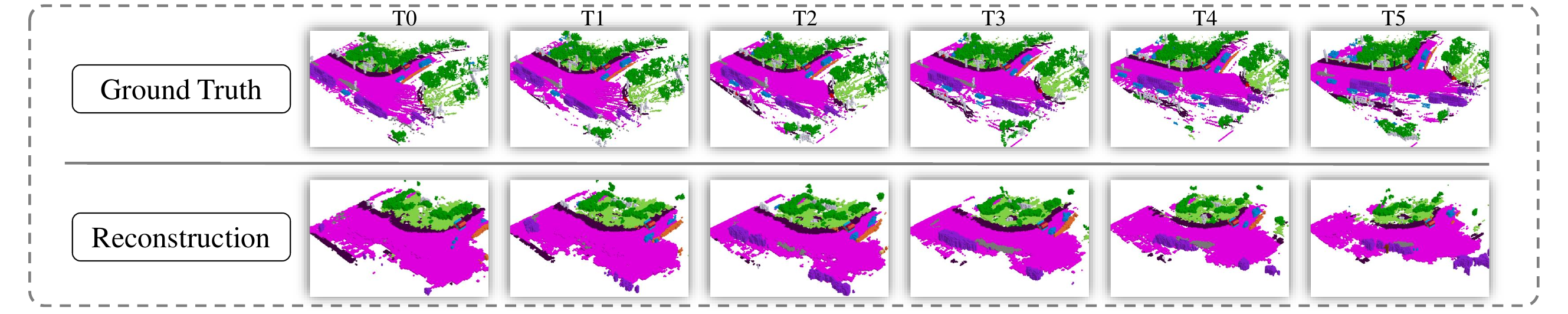}
\vspace{-5mm} 
  \caption{\textbf{Visualization of reconstruction of the 4D occupancy scene tokenizer.}}
  \label{exp1}
\vspace{-5mm} 
\end{figure}

\begin{table}[t!]
\small
  \centering
\setlength{\tabcolsep}{18pt}
  \caption{\textbf{Comparisons of OccSora with other models in its generation capability.} To the best of our knowledge, we are the first 4D occupancy generation model. Therefore, we only provide comparisons with other generative models on different datasets and data formats.
  }
   \begin{tabular}{@{}cccccc@{}}
    \toprule
    Method & Type & Dimension & Dataset &  FID \\
    \midrule
    DiT \cite{peebles2023scalable} & Image & 2D & ImageNet \cite{deng2009imagenet} & 12.03 \\
    MagicDriver \cite{gao2023magicdrive} & Video & 3D & nuScenes \cite{caesar2020nuscenes} & 14.46 \\
    DriveDreamer \cite{wang2023drivedreamer} & Video & 3D & nuScenes \cite{caesar2020nuscenes} & 14.9 \\
    DriveGAN \cite{Kim_2021_CVPR} & Video & 3D & nuScenes \cite{caesar2020nuscenes}  & 27.8 \\
    SemCity \cite{lee2024semcity} & Occupancy & 3D & KITTI \cite{geiger2012we} & 40.63 \\
    OccSora & Occupancy Video & 4D & nuScenes \cite{caesar2020nuscenes} & \textbf{8.348} \\

    \bottomrule
    \end{tabular}%
\vspace{-3mm}
  \label{tab4}%
\end{table}%

\begin{figure}[t!]
  \centering
  \includegraphics[width=1\linewidth]{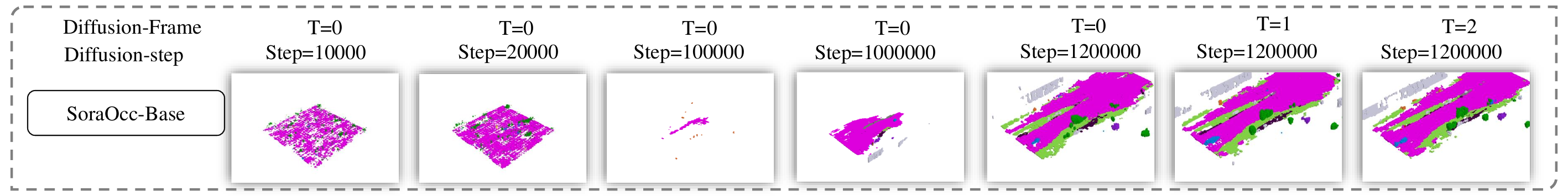}
\vspace{-6mm}
  \caption{\textbf{The visualization of the progressive generation of accurate scenes as the model undergoes iterative training.} }
  \label{exp4}
\vspace{-5mm}
\end{figure}

\subsection{4D Occupancy Reconstruction}

The compression and reconstruction of 4D occupancy are essential for learning the latent spatiotemporal correlations and features necessary for image generation. Unlike traditional models for video and image processing, OccSora operates one dimension higher than occupancy for single frames and two dimensions higher than images.  Therefore, achieving efficient compression and accurate reconstruction is paramount. Figure \ref{exp1} depicts the ground truth and reconstruction of the occupancy. We also conducted a quantitative analysis of 4D occupancy reconstruction, as shown in Table \ref{tab1}. The table indicates that even with OccSora achieving a compression ratio 32 times greater than that of OccWorld, it still maintains nearly 50\% mIoU of the original OccWorld model. This unified temporal compression effectively captures the dynamic changes of various elements, improving long-sequence modeling capabilities compared to progressive autoregressive methods.

\subsection{4D Occupancy Generation}

In the diffusion-based world model for the 4D occupancy generation task, we used tokens generated by the OccSora model, trained with 32 frames, as input for our generation experiments. In Figure \ref{exp4}, we present the visual results of  across training iterations, from 10,000 to 1,200,000 steps. These visual results indicate that as the number of training iterations increases, the accuracy of the OccSora model continuously improves, demonstrating the generation of coherent scenes.

\begin{figure}[t!]
  \centering
  \includegraphics[width=1\linewidth]{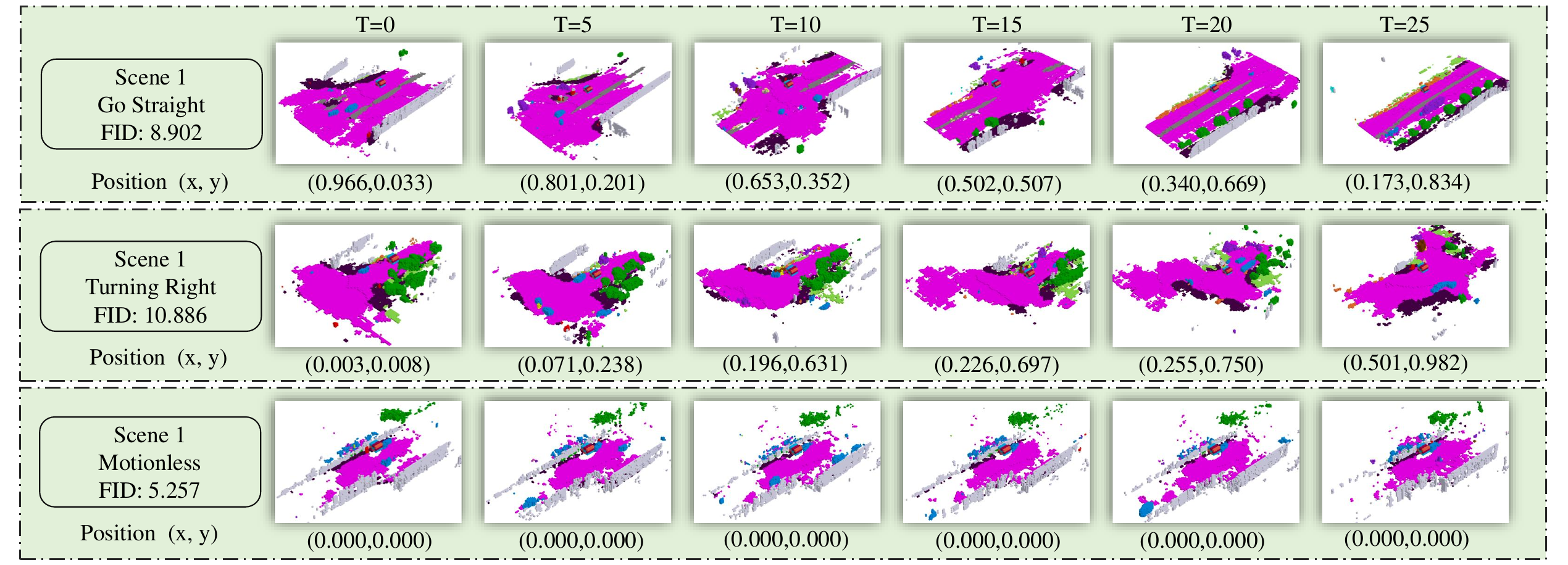}
\vspace{-5mm}
  \caption{\textbf{4D occupancy generation under different input trajectories.} From top to bottom, there is go straight, turning right, and motionless, with each scene generation corresponding to the trajectory, ensuring logical coherence and continuity.}
  \label{exp5}
\vspace{-5mm}
\end{figure}

\begin{figure}[t!]
  \centering
  \includegraphics[width=1\linewidth]{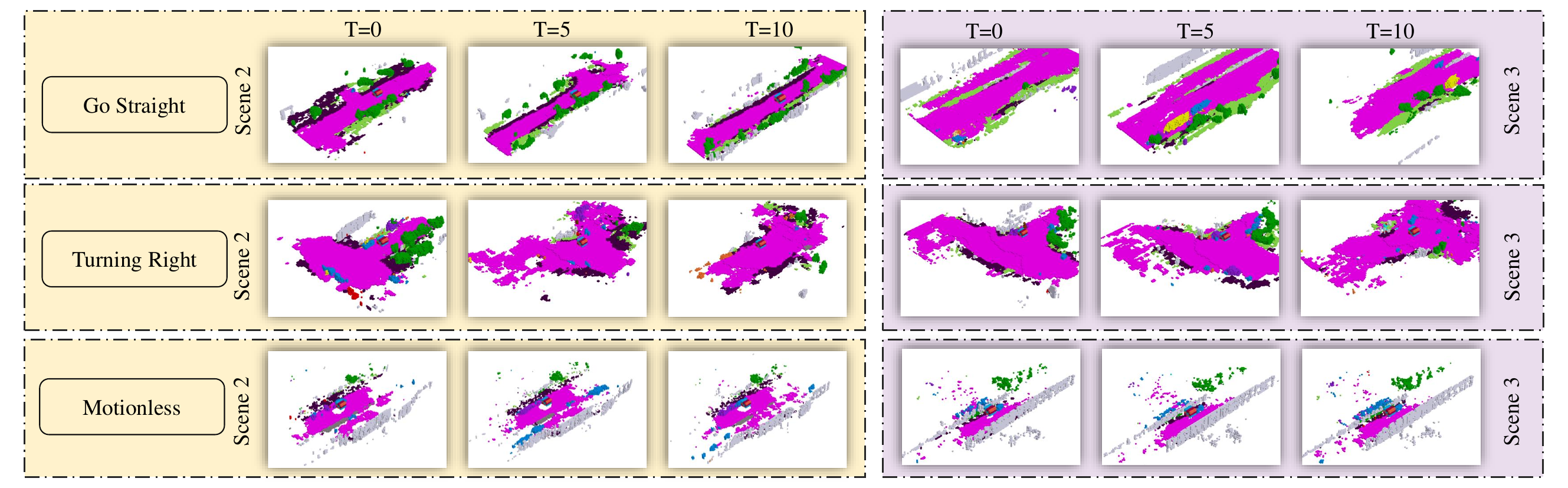}
\vspace{-5mm}
  \caption{\textbf{Generating diverse continuous scenes under trajectory control.} The generated scenes exhibit diversity while maintaining the stability of the original trajectory control.}
  \label{exp6}
\vspace{-5mm}
\end{figure}

\begin{table}[t!]
\small 
  \centering
  \caption{\textbf{Results of ablative evaluation on different components.} We quantitatively evaluated the impact of different compression rates, components, and channel dimensions on the reconstruction and generation results through controlled variables.}
      \begin{tabular}{@{}ccccccccccc@{}}
    \toprule
    Input Size \( R_{in} \) & Token Size $R_{mi}$ & Channel & Class & T embed. & Trajectory & IoU & mIoU & FID \\
    \midrule
    32x200x200 & 128x4x25x25 & 8 & \cmark & \cmark & \cmark & \textbf{37.03} & \textbf{27.42} & \textbf{8.34} \\
    32x200x200 & 128x4x25x25 & 8 & \cmark & \xmark & \cmark & \textbf{37.03} & \textbf{27.42} & 87.26 \\
    32x200x200 & 128x4x25x25 & 8 & \cmark & \cmark & \xmark & \textbf{37.03} & \textbf{27.42} & 17.48 \\
    32x200x200 & 128x4x25x25 & 4 & \cmark & \cmark & \cmark & 29.67 & 23.21 & 34.24 \\
    32x200x200 & 128x8x50x50 & 8 & \cmark & \cmark & \cmark & 32.91	
 & 24.4 & 72.32 \\
 \cline{0-8}
    12x200x200 & 64x3x50x50 & 8 & \cmark & \cmark & \cmark & 26.73	
 & 14.12 & 187.78 \\
    12x200x200 & 64x3x25x25 & 8 & \cmark & \cmark & \cmark & 22.42 & 9.27 & 270.23 \\
    12x200x200 & 32x3x25x25 & 8 & \cmark & \cmark & \cmark & 13.60 & 3.85 & 465.18 \\
    \bottomrule
    \end{tabular}%
  \label{tab7}%
\vspace{-7mm}
\end{table}%

We compared and quantitatively evaluated our proposed OccSora model against other generation models. As the first 4D occupancy world model for autonomous driving, we only compared it against conventional image generation, 2D video generation, and static 3D occupancy scene generation methods. As shown in Table \ref{tab4}, our model achieves similar performance in terms of the Fréchet Inception Distance (FID) \cite{heusel2018gans}, demonstrating the effectiveness of the proposed method.

\textbf{Trajectory Video Generation.} OccSora has the capability to generate various dynamic scenes based on different input trajectories, thus learning the relationship between ego vehicle trajectories and scene evolution in autonomous driving. As shown in Figure \ref{exp5}, we input different vehicle trajectory motion patterns into the model, demonstrating the 4D occupancy for go straight, turning right, and motionless. We conducted experiments at different scales for generating trajectories, revealing that the FID score is lowest for stationary scenes and higher for curved scene, indicating the complexity of continuously modeling curved motion scenes and the simplicity of modeling stationary scenes.

\begin{table}[t]
\small 
  \centering
  \caption{\textbf{The quantitative analysis of different scales regarding denoising steps and denoising rates.} Denoising steps have a relatively minor impact on the model, whereas denoising rates and model scales significantly affect the quality of the generated outputs.}
\setlength{\tabcolsep}{8pt}
\begin{tabular}{@{}ccc|ccccccc@{}}
    \toprule
    \multirow{2}{*}{Step} & \multirow{2}{*}{Input Size $R_{in}$} & \multirow{2}{*}{Token Size $R_{mi}$} & \multicolumn{6}{c}{FID} \\
    \cmidrule(lr){4-9}
     & & & 10\% & 30\% & 50\% & 70\% & 90\% & 100\% \\
    \midrule
    10 & 32x200x200 & 128x4x25x25 & 49863 & 34927 & 17630 & 339 & 42 & 9.1 \\
    100 & 32x200x200 & 128x4x25x25 & 53297 & 29521 & 19471 & 1084 & 72 & 10.08 \\
    1000 & 32x200x200 & 128x4x25x25 & 32171 & 10284 & 5924 & 591 & 17 & 8.94 \\
    \midrule
    10 & 12x200x200 & 64x3x50x50 & 71293 & 54625 & 5644 & 7416 & 742 & 431 \\
    100 & 12x200x200 & 64x3x50x50 & 81274 & 53431 & 45346 & 3161 & 456 & 446 \\
    1000 & 12x200x200 & 64x3x50x50 & 43631 & 33415 & 17431 & 4366 & 379 & 353 \\
    \bottomrule
\end{tabular}
  \label{tab8}%
\vspace{-4mm}
\end{table}%

\begin{figure}[t]
  \centering
  \includegraphics[width=1\linewidth]{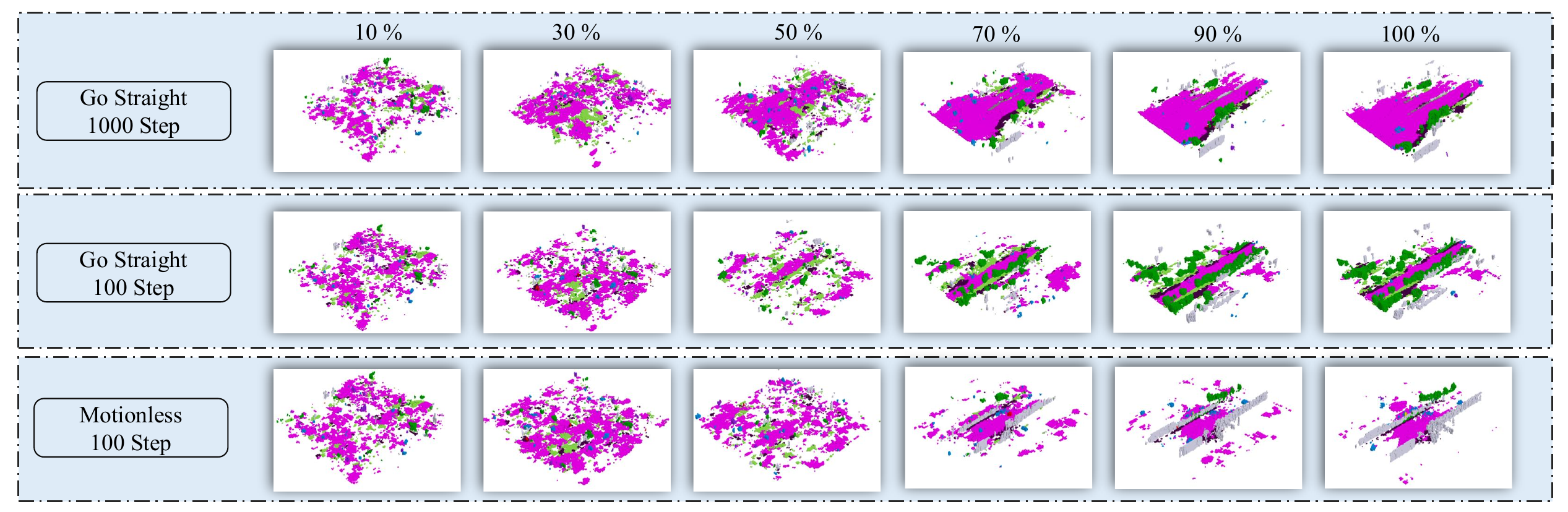}
\vspace{-7mm}
  \caption{\textbf{Effect of the denoising ratios under different trajectories or denoising steps.} Denoising steps and trajectories have minor impacts while denoising ratios have a significant effect.}
  \label{exp7}
\vspace{-7mm}
\end{figure}

\textbf{Scene Video Generation.} Diversity in scenes is crucial under reasonable trajectory control. We tested the reconstruction of 4D occupancy scenes for different scenarios under three trajectories to verify the generalization performance of generating scenes under controllable trajectories. In Figure \ref{exp6}, the left and right parts respectively demonstrate the capability to generate different scenes under the same trajectory. In the reconstructed scenes, surrounding trees and road environments exhibit random variations while still maintaining the logic of the original trajectory, showcasing the ability to maintain robustness in generating scenes corresponding to the original trajectory and its generalization across different scenarios.

\subsection{Ablation and Analysis}

\textbf{ Analysis of the Tokenizer and Embeddings.} 
We conducted an ablation of the proposed components including different compression scales, the number of class tokenizer discretizations, time-step embeddings, and vehicle trajectory embeddings, as shown in Table \ref{tab7}. 
When the number of class tokenizer discretizations was reduced from 8 to 4, the reconstruction accuracy dropped by approximately 18\%. The FID score also declined after removing the time-step embeddings. Without position embeddings, the generated scenes lacked motion control and displayed almost linear movement patterns influenced by the data distribution. Additionally, at lower compression ratios, although the reconstruction performance was better compared to higher compression ratios, the lack of higher-dimensional feature correlations prevented the generation of effective scenes.

\textbf{Analysis of the Generation Steps.} The total number of denoising steps and the denoising rate can affect the generation quality to some extent. As shown in Figure \ref{exp7}, as the denoising rate increases, the generated scenes become progressively clearer. According to the quantitative results in Table \ref{tab8}, increasing the total number of denoising steps can improve generation accuracy to a certain extent. However, the generation quality is much more significantly influenced by the token size and the number of channels than by the total number of denoising steps.

\section{Conclusion and Limitations}

In this paper, we have introduced a framework for generating 4D occupancy to simulate 3D world development in autonomous driving. Using a 4D scene tokenizer, we obtain compact representations for input and achieve high-quality reconstruction for long-sequence occupancy videos. Then, we learn a diffusion transformer on the spatiotemporal representations and generate 4D occupancy conditioned on a trajectory prompt. Through experiments on the nuScenes dataset, we demonstrate accurate scene evolution. In the future, we will investigate more refined 4D occupancy world models and explore the possibilities of end-to-end autonomous driving under closed-loop settings.

\textbf{Limitations.} The advantage of a 4D occupancy world model lies in establishing an understanding of the relationship between scenes and motion. However, due to limitations in the granularity of voxel data, we cannot construct more finely detailed 4D scenes. 
The generative results also demonstrate inconsistent details for moving objects, possibly due to the small size of the training data.

\newpage
{
\small
\bibliographystyle{plain}
\bibliography{main}
}

\newpage

\appendix

\section{Appendix}

\subsection{More Visualizations}
\begin{figure}[h]
  \centering
  \includegraphics[width=0.92\linewidth]{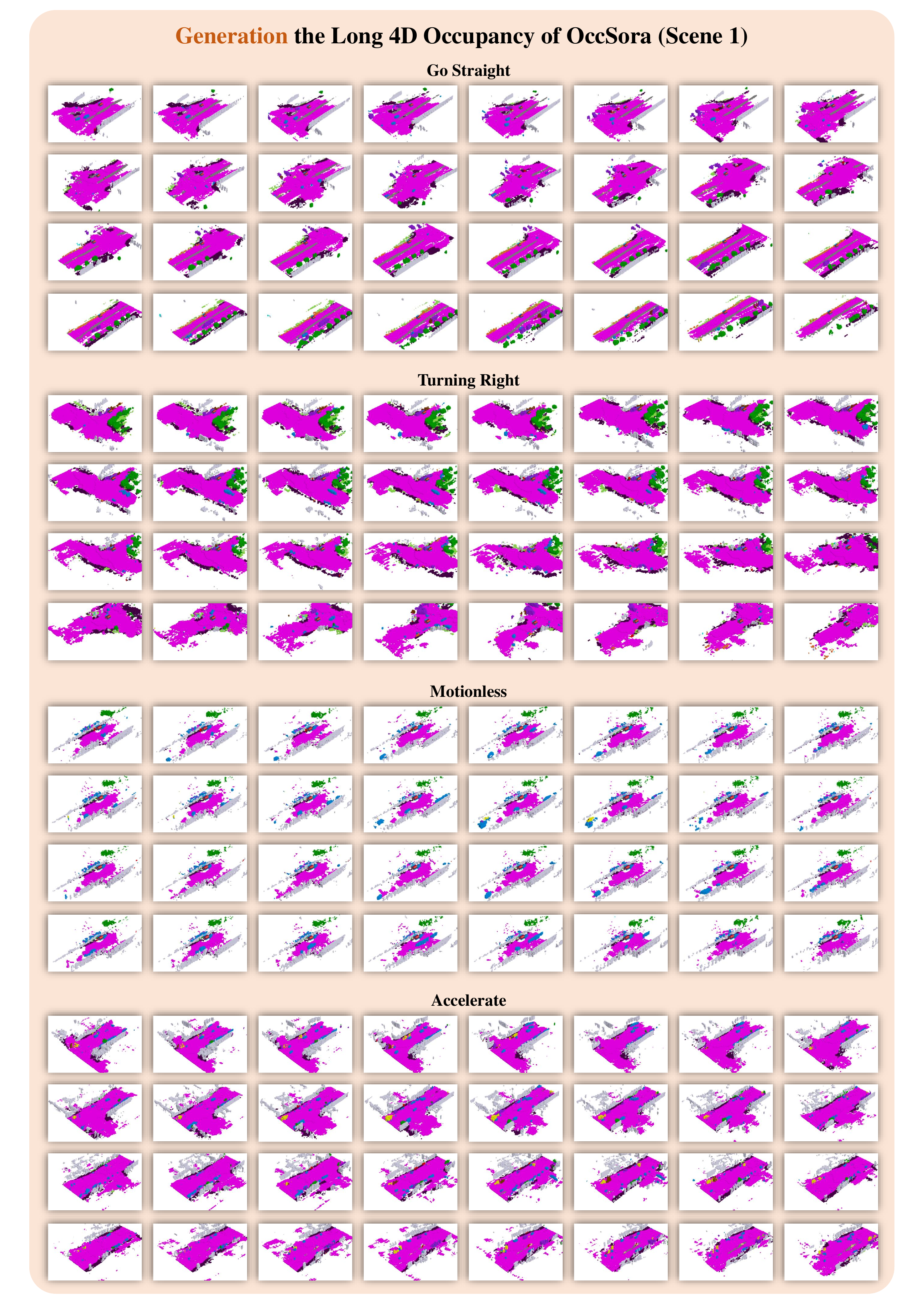}
  \caption{\textbf{The ability to generate long time-series 4D occupancy under different trajectory controls.} From top to bottom, we present long-term continuous scenes generated under four types of ego vehicle trajectories: Go Straight, Turning Right, Motionless, and Accelerate.}
  \label{app01}
\end{figure}

\begin{figure}[h]
  \centering
  \includegraphics[width=0.92\linewidth]{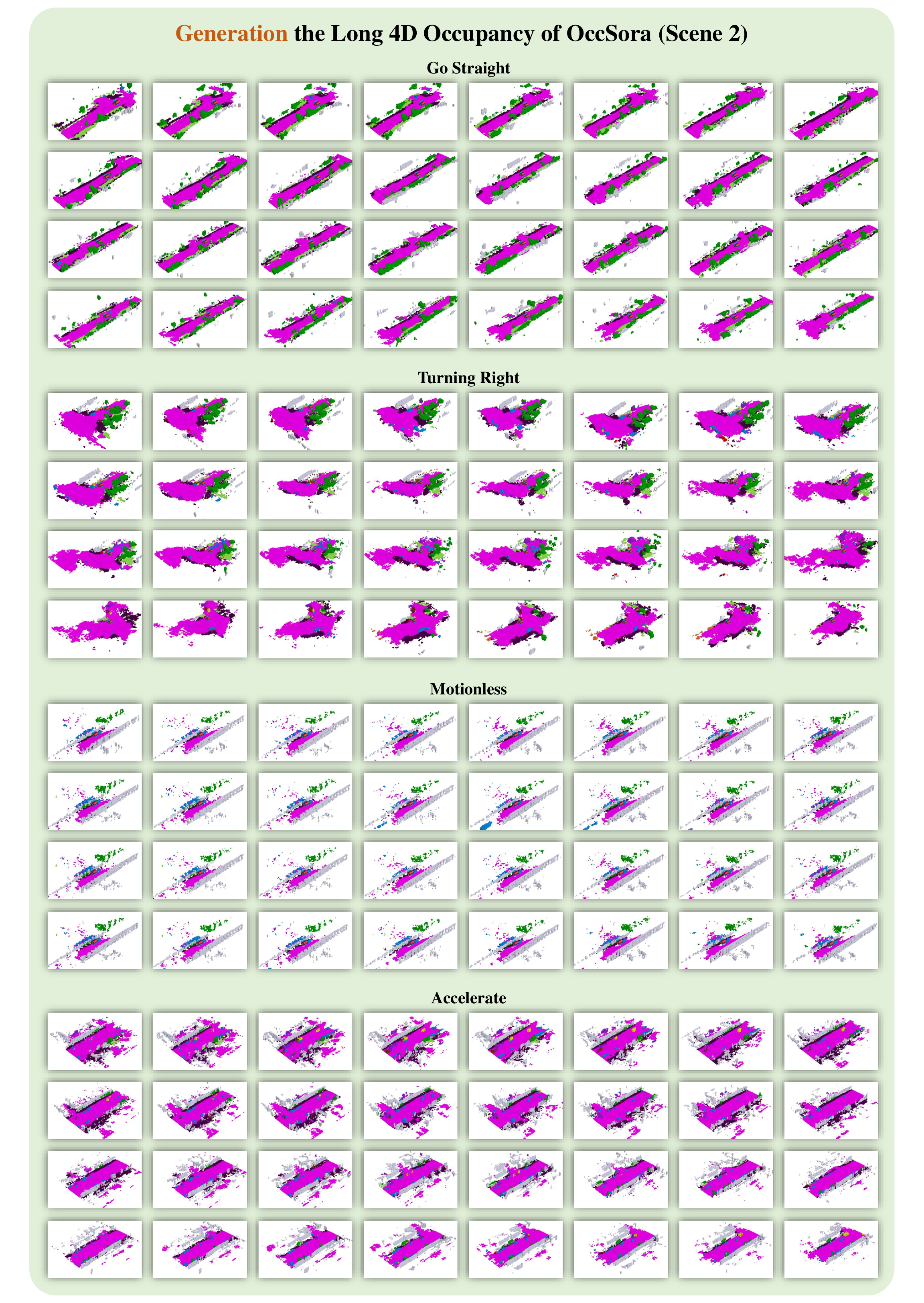}
  \caption{\textbf{The generalization ability to generate different scenes under fixed ego vehicle trajectories.} From top to bottom, we show the capabilities of generating different scenes under the four vehicle trajectories: Go Straight, Turning Right, Motionless, and Accelerate.}
  \label{app02}
\end{figure}

To provide results for longer time series, we present the generated scenes under different ego vehicle trajectory controls, namely Go Straight, Turning Right, Motionless, and Accelerate in Figure \ref{app01}. Additionally, we also showcase the equivalent control methods under different scenes in Figure \ref{app02}.

\newpage

\end{document}